\documentclass[10pt,twocolumn,letterpaper]{article}
\usepackage[accsupp]{axessibility} 
 \usepackage{cvpr}              

\usepackage[dvipsnames]{xcolor}

\definecolor{cvprblue}{rgb}{0.21,0.49,0.74}

\def\paperID{*****} 
\def\confName{CVPR}
\def\confYear{2024}

\title{MoPEFT: A Mixture-of-PEFTs for the Segment Anything Model}

\author{Rajat Sahay\\
Rochester Institute of Technology\\
Rochester, NY USA\\
{\tt\small rs6287@rit.edu}
\and
Andreas Savakis\\
Rochester Institute of Technology\\
Rochester, NY USA\\
{\tt\small andreas.savakis@rit.edu}
}

\begin{document}
\maketitle
\begin{abstract}
The emergence of foundation models, such as the Segment Anything Model (SAM), has sparked interest in Parameter-Efficient Fine-Tuning (PEFT) methods that tailor these large models to application domains outside their training data. However, different PEFT techniques modify the representation of a model differently, making it a non-trivial task to select the most appropriate method for the domain of interest. We propose a new framework, Mixture-of-PEFTs methods (MoPEFT), that is inspired by traditional Mixture-of-Experts (MoE) methodologies and is utilized for fine-tuning SAM. Our MoPEFT framework incorporates three different PEFT techniques as submodules and dynamically learns to activate the ones that are best suited for a given data-task setup. We test our method on the Segment Anything Model and show that MoPEFT consistently outperforms other fine-tuning methods on the MESS benchmark.
\end{abstract}    
\section{Introduction}
\label{sec:intro}

The machine learning research community has witnessed an explosion in the development of foundation models in recent years, such as CLIP \citep{radford2021learning}, GPT-4 \citep{achiam2023gpt}, and PaLM \citep{chowdhery2023palm}. More recently, the Segment Anything Model (SAM) \citep{kirillov2023segment}, a promptable model pretrained on over 1 billion masks and 11 million images, emerged as a foundation model for image segmentation. It has demonstrated performance comparable to state-of-the-art approaches in multiple applications related to segmentation tasks. Moreover, SAM's zero-shot and few-shot capabilities 
have garnered significant attention across multiple domains \citep{yang2023track,ke2024segment}.
However, prior works \citep{mazurowski2023segment,osco2023segment} have shown that despite noteworthy proficiency in segmenting real-world objects in natural images, SAM has difficulty with objects outside its training domain.

Following the pretraining-fine-tuning paradigm \citep{xu2021rethinking}, it is desirable to fine-tune SAM 
in order to enhance its performance in the application domain of interest.  However, fine-tuning foundation models can be costly due to their large number of parameters. This motivates the development of fine-tuning methods
with the goal of
achieving comparable performance to full fine-tuning while employing as few trainable parameters as possible. 
Interest in
Parameter-Efficient Fine-Tuning methods (PEFT) 
has increased significantly
since the advent of foundation models \citep{hu2021lora,lester2021prompt,jia2022vpt,pfeiffer2020adapterhub}.

Recent studies \citep{he2021adapter,li2021prefix} have shown that some PEFT methods are more effective at fine-tuning with the objective of reducing overfitting on the target domain- especially in data-sparse environments. However, we find that combining different PEFT methods often yields better results without a substantial loss in efficiency. This is because different techniques operate on different parts of the transformer architecture, making it possible to use more than one technique at a time.

In light of this, we propose a new framework, called Mixture-of-PEFTs (MoPEFT), that incorporates different PEFT methods as submodules and learns to dynamically activate the fine-tuning method(s) that best suit the data or task of interest. 
Inspired by the Mixture-of-Experts approach \citep{nguyen2018practical,lin2024moe,pieri2024bimedix}, MoPEFT
switches between different PEFT methods using a \textit{gating mechanism} that learns to favor 
the method that positively contributes to a given task. 
In addition, since the number of parameters introduced by each PEFT is very small, e.g. compared to the entire SAM architecture, combining multiple PEFT methods has little effect on the efficiency of our framework. In this paper, we include the three most commonly used PEFT techniques- LoRA \citep{hu2021lora}, Prefix Tuning \citep{jia2022vpt}, and Adapters \citep{he2021adapter}. Our experiments shed light on the effectiveness of these methods across multiple domains, and their effectiveness when combined together in our MoPEFT framework.

Our contributions can be outlined as follows: (i) We conduct a comprehensive survey of the widely-used PEFT methods and benchmark their performance across multiple domains; (ii) We introduce our MoPEFT framework, which incorporates multiple PEFT methods as submodules and learns to dynamically activate or deactivate the appropriate submodule based on the given task; and (iii) We show that our MoPEFT framework achieves better performance than  individual PEFT methods across multiple domains in the MESS benchmark.


\begin{figure}
    \centering
    \includegraphics[width=0.7\linewidth]{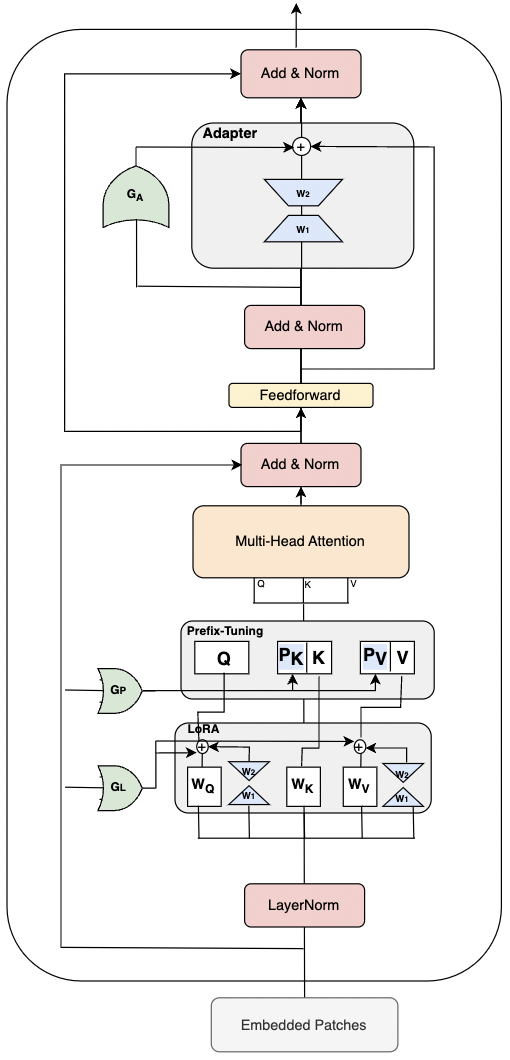}
    \caption{An overview of our MoPEFTs framework}
    \label{fig:mopefts}
\end{figure}


\section{Mixture-of-PEFTs (MoPEFT)}

In this section, 
we present the details of
our proposed MoPEFT framework, illustrated in Figure \ref{fig:mopefts}.

\subsection{Parameter Efficient Fine-Tuning Methods}
\noindent {\bf Low Rank Adaptation (LoRA).}
Low Rank Adaptation (LoRA) \citep{hu2021lora} exploits the \textit{low rank} structure inherent in deep learning models to align them to specific tasks. It works by introducing trainable low-rank matrices and combines them with the original matrices in the multi-head self-attention (MHSA) blocks. The pre-trained weight matrix $W_{0} \in \mathbb{R}^{d \times k}$ is updated as $W_{0} + \Delta W$, where $\Delta W \in \mathbb{R}^{d \times k}$ is a low-rank matrix decomposed as $\Delta W = BA$. Here, $B \in \mathbb{R}^{d \times r}, A \in \mathbb{R}^{r \times k}$ and the rank $r << min(d,k)$. During fine-tuning, the pre-trained weights remain frozen, and $\Delta W$ serves as the trainable parameter. The decomposition of $\Delta W = BA$ as a product of two low-rank matrices effectively reduces the memory and computational cost of fine-tuning.

\vspace{4pt}
\noindent {\bf Prefix Tuning.}
Prefix Tuning \citep{li2021prefix} prepends a number of tunable, task-specific vectors to the input of the multi-head self-attention in \textit{each} Transformer block, which original tokens can attend to as if they were virtual tokens. This method was originally developed for natural language processing and was eventually extended to vision applications as Deep Visual Prompt Tuning (VPT-Deep) \citep{jia2022vpt}. We use VPT-Deep for all our experiments and call it Prefix Tuning to maintain uniformity with literature in the field. We denote the original sequence length $L_0$, the number of tunable vectors (i.e., prefix length) as $L$, and the Transformer layer input as $h_{in} \in \mathbb{R}^{{D_{hidden} \times L_0}}$. First, three linear projections, $W_Q, W_K, W_V \in \mathbb{R}^{D_{hidden} \times D_{hidden}}$ transform $h_{in}$ into Query $(Q)$, Key $(K)$, and Value $(V)$ matrices. The two prefix matrices $P_K \in \mathbb{R}^{D_{hidden} \times L}$ and $P_V \in \mathbb{R}^{D_{hidden} \times L}$ are pre-pended to $K$ and $V$. The prefix matrix $P$ is reparametrized by a feedforward network to stabilize the optimization procedure.

\vspace{4pt}
\noindent 
{\bf Adapters.}
Adapters \citep{he2021adapter} align the model to the target task by adding a trainable MLP after the feedforward layer in each Transformer block. The MLP consists of a down+up projection that condenses and recovers the size of the original hidden token space. Mathematically, we can denote the Adapter operation as 
\begin{equation}
   Z_{A} = W_{1}^{T} \phi (W_{2}^{T}Z_{FN})
\end{equation}
where, $W_1 \in \mathbb{R}^{D_{hidden} \times D_{mid}}, W_2 \in \mathbb{R}^{D_{mid} \times D_{hidden}}$. Here, $D_{hidden}$ represents the hidden token space in the Transformer block, and $D_{mid}$ represents the condensed embedding space of the Adapter MLP. $Z_{FN}$ is the output of the feedforward network of the Transformer block after the residual connection and the layer normalization steps.


\subsection{Task Formulation}

Given a very large model $M$, which cannot be efficiently fine-tuned due to computational costs, assume we have a selection of PEFT methods $FT \in [ft_1, ft_2...ft_n]$, each of which have negligible trainable parameters compared to $M$ (ie.$ \sum_{n} FT << |M|$). Our goal is to design a framework that incorporates $[ft_1, ft_2...ft_n]$ as individual, independent submodules and learns to dynamically activate different $ft_i$ based on different data-task scenarios. This would ensure that a singular framework would be capable of achieving optimal results in terms of both accuracy and efficiency without permuting through all data-task combinations for every datapoint.

\subsection{Proposed Method}

\paragraph{Intuition.}
During the analysis of individual PEFT methods, we observed that different methods often involve different parts of the Vision Transformer model in the image encoder of SAM. For instance, Adapters add an MLP after the feedforward layer in each Transformer block, while Prefix Tuning prepends tunable tensors before the multi-head self-attention layers. This unique property makes it possible to essentially combine multiple PEFT techniques in the proposed framework without interfering with each other. 

Keeping the above in mind, we propose a unified MoPEFT framework which takes a hybrid approach by incorporating multiple PEFT methods as submodules. At a high level, MoPEFT shows better performance than its individual components due to two main reasons. Firstly, MoPEFT learns to dynamically 
access individual
submodules based on the given task. This means that for a given data-task sample, a particular PEFT method may be allotted different weights or turned off entirely to ensure optimal performance in all cases. Secondly, we find that our MoPEFT framework generally outperforms the best-performing individual PEFT technique in multiple domains, suggesting that there may be benefits due to compounding effects that lead to better model effectiveness, as multiple PEFT techniques are used together. We show how we incorporate these different techniques under one framework in the next section.

\vspace{4pt}
\noindent {\bf Gating Mechanism.}
To achieve fine-grained control over the activation of the individual PEFT techniques that make up our MoPEFT framework, we take inspiration from current Mixture-of-Experts (MoE) methods \citep{lin2024moe, wang2022adamix}. Similar to the Sparsely-Gated-MoE method \citep{shazeer2017outrageously}, we add a gating mechanism that dynamically links different PEFT methods to the relevant layers in the image encoder of SAM. As depicted in Figure~\ref{fig:mopefts}, we add three trainable gates, one for each PEFT technique. Intuitively, if a particular PEFT technique is useful for a given data-task setup, then the output of the corresponding gate would be set to high. This would ensure that the specific PEFT plays a more important role during the execution. 

For LoRA, our gate is not added directly in the form of the traditional MLP architecture as seen in MoE literature. Instead, we make use of the inherent \textit{scaling factor}, $\alpha$ already present in the LoRA architecture as a pseudo-gating mechanism. A higher $\alpha$ assigns more weight to the LoRA activations, while a lower $\alpha$ makes the effect of LoRA negligible. Thus, we already have a gating mechanism in place. To integrate this with our broader framework, we make the scaling factor learnable by using a feedforward network instead of specifying the constant manually.

For Prefix Tuning, we design a gating function $G_P \in (0,1)$ that is applied to the Prefix vectors $P_K$ and $P_V$ keeping the representations of the original Key and Value tokens $K$ and $V$ intact. $G_P$ is estimated using another feedforward network which takes in the input provided to the specific ViT layer.

For Adapters, we make use of the residual connection between the Adapter MLP and the feedforward network of the ViT Transformer block. This connection is responsible for summing up the input to the Adapter MLP. Our Adapter Gating Function $G_A \in (0,1)$ estimates the importance of the Adapter MLP using a feedforward network with sigmoid activation. The Adapter MLP is essentially bypassed if $G_A = 0$.
\section{Experiments}

\begin{figure*}[!ht]
    \centering
    \begin{subfigure}[t]{0.3\linewidth}
        \centering
        \includegraphics[height=1.2in]{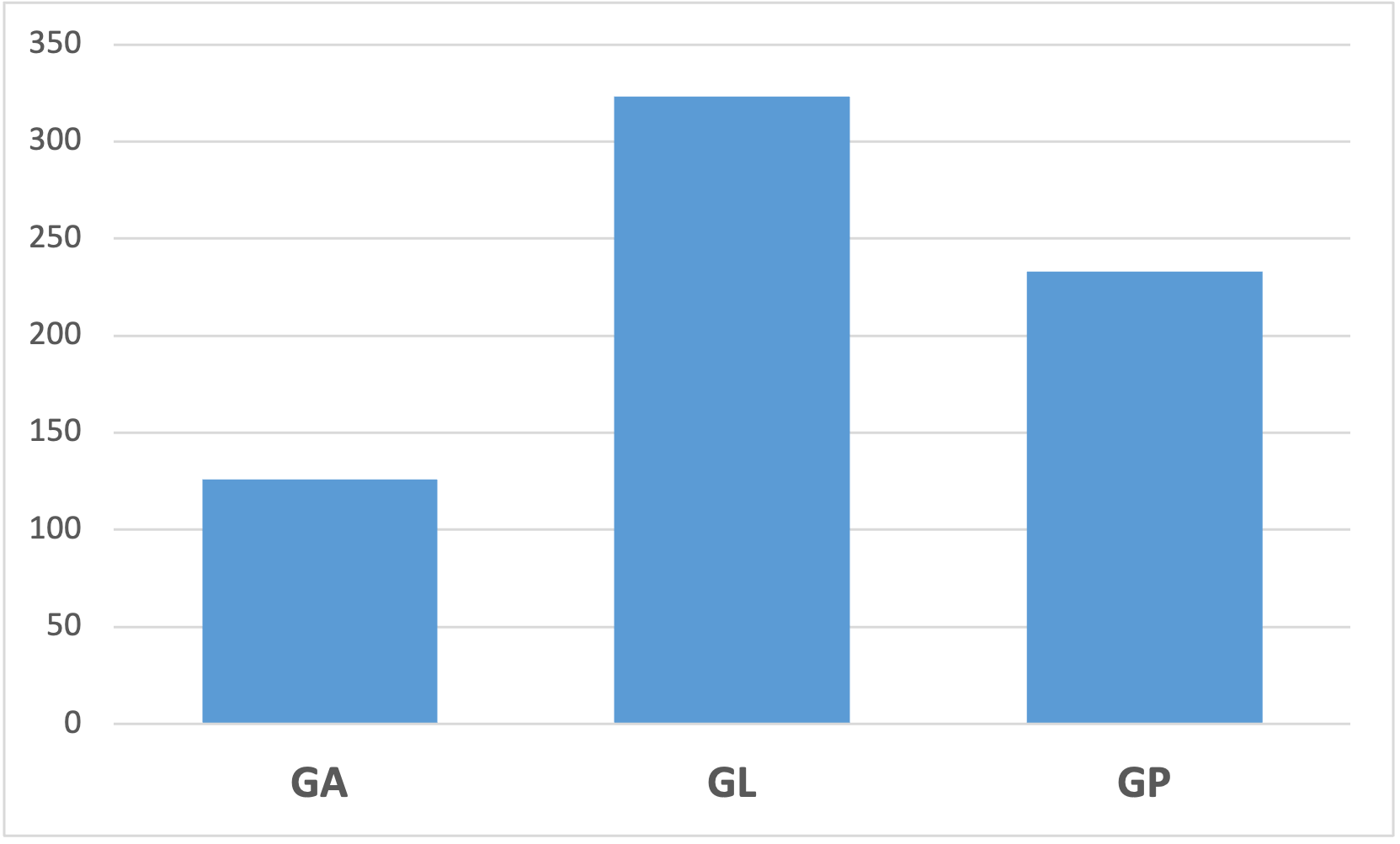}
        \caption{iSAID \citep{waqas2019isaid}}
    \end{subfigure}%
    ~ 
    \begin{subfigure}[t]{0.3\linewidth}
        \centering
        \includegraphics[height=1.2in]{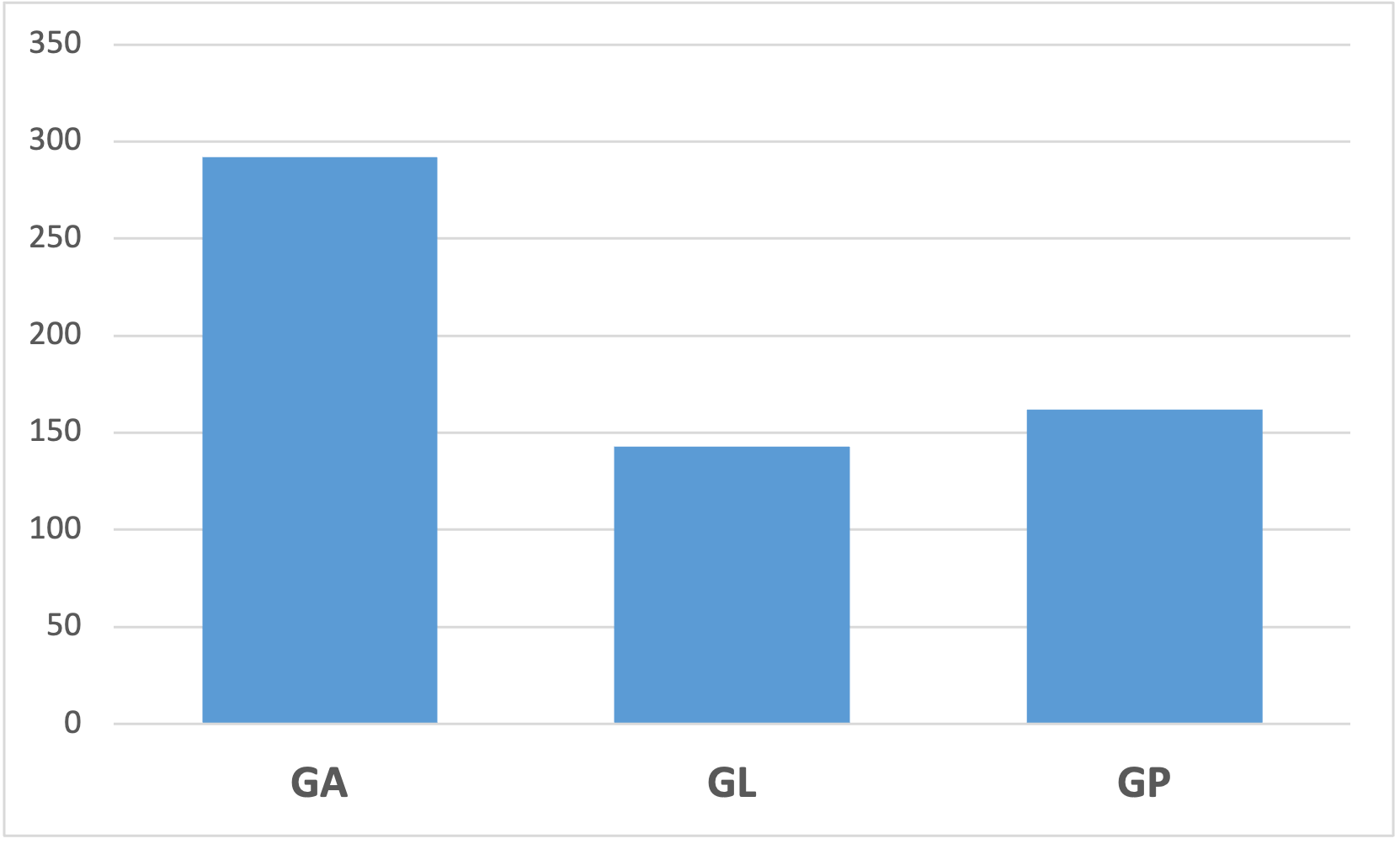}
        \caption{Kvasir-Instrument \citep{jha2021kvasir}}
    \end{subfigure}
    \caption{Number of times each PEFT method is called during inference. Different datasets display distinct patterns}
    \label{fig:graphs}
\end{figure*}


\begin{table*}[t]
    \centering
    \begin{tabular}{ | l | l | c | c | c | c | c | c | c | }
\hline
	\textbf{Domain} & \textbf{Dataset} & \textbf{Baseline} & \textbf{decoderFT} & \textbf{LoRA} & \textbf{VPT Deep} & \textbf{VPT} & \textbf{Adapter} & \textbf{MoPEFTs} \\ \hline
	\textbf{General} & BDD100K \citep{yu2020bdd100k} & 41.58 & 42.84 & 49.39 & 46.24 & 43.18 & 47.03 & 50.93 \\ 
	 & Dark Zurich \citep{darkzurich} & 20.91 & 23.42 & 30.82 & 27.16 & 24.49 & 26.72 & 31.19 \\ 
	 & MHPv1 \citep{mhpv1} & 29.38 & 31.40 & 39.21 & 35.59 & 32.72 & 36.17 & 41.84 \\ 
	 & FoodSeg103 \citep{foodseg} & 10.48 & 14.93 & 22.45 & 19.91 & 16.02 & 20.05 & 22.99 \\ 
	 & ATLANTIS \citep{atlantis} & 17.33 & 20.62 & 28.03 & 27.61 & 24.91 & 27.91 & 30.03 \\ 
	 & DRAM \citep{dram} & 57.38 & 58.83 & 64.48 & 60.23 & 58.89 & 63.79 & 67.25 \\ \hline
	\textbf{Earth } & iSAID \citep{waqas2019isaid} & 62.59 & 63.14 & 66.29 & 65.61 & 64.71 & 64.82 & 68.29 \\ 
	 \textbf{Monitoring} & ISPRS Postdam \citep{isprspostdam} & 29.73 & 29.92 & 38.25 & 33.52 & 31.42 & 35.77 & 40.42 \\ 
	 & WorldFloods \citep{worldfloods} & 46.45 & 48.75 & 59.53 & 56.13 & 52.29 & 54.23 & 63.17 \\ 
	 & FloodNet \citep{rahnemoonfar2021floodnet} & 39.72 & 40.94 & 46.79 & 43.97 & 42.81 & 41.09 & 50.01 \\ 
	 & UAVid \citep{lyu2020uavid} & 60.19 & 60.96 & 69.43 & 65.39 & 61.19 & 63.59 & 71.12 \\ \hline
	\textbf{Medical} & Kvasir-Instr. \citep{jha2021kvasir} & 46.82 & 48.32 & 66.97 & 58.31 & 52.23 & 62.06 & 71.92 \\ 
    \textbf{Imaging}	 & CHASE DB1 \citep{chasedb1} & 23.56 & 25.95 & 37.22 & 32.48 & 28.39 & 30.85 & 42.49 \\ 
	 & CryoNuSeg \citep{mahbod2021cryonuseg} & 38.06 & 40.36 & 54.93 & 48.12 & 44.81 & 36.22 & 59.88 \\ 
	 & PAXRay-4 \citep{paxray} & 41.07 & 43.73 & 56.05 & 52.83 & 46.62 & 51.35 & 59.42 \\ \hline
  \end{tabular}
  \caption{Comparison of our MoPEFTs framework with fine-tuned SAM variants across multiple domains. Scores shown are mIOU scores.}
  \label{tab:all_main}
\end{table*}


\vspace{4pt}
\noindent {\bf Datasets}.
We employ the Multi-domain Evaluation of Semantic Segmentation (MESS) benchmark \citep{blumenstiel2024mess}, which measures the mIOU score of models performing semantic segmentation tasks on 22 datasets spread across five major domains- General, Earth Monitoring, Medical Imaging, Engineering, and Agriculture and Biology. For brevity, we present results on only the first three domains. The datasets are not distributed evenly across all domains (for instance, General has six datasets while Engineering has three) but we examine at performance on individual dataset as opposed to collective domains.  

\vspace{4pt}
\noindent {\bf Implementation Details.}
We use the Segment Anything Model \citep{kirillov2023segment} for all our fine-tuning and experiments. The traditional implementation of SAM consists of an image encoder (we use ViT-B for our experiments), a Prompt Encoder and a Mask Decoder.  However, to better equip SAM for end-to-end semantic segmentation, we freeze the Prompt Encoder, always providing constant prompt tokens to the Mask Decoder when fine-tuning. Additionally, we apply full fine-tuning to the Mask Decoder, since it is an extremely lightweight module.

For consistency, we include public implementations for all PEFT methods in our framework. We use a batch size of 4 and the Adam optimizer with a learning rate of 1x10$^{-4}$ as a default with a weight decay of 1x10$^{-4}$. All PEFT methods are implemented in the same codebase to ensure a fair comparison. We largely follow the default PEFT-specific hyperparameters and keep them unchanged across domains for uniformity. Unless otherwise specified, we set the LoRA rank $r=8$ prefix length $L=20$, and the adapter bottleneck size $D_{mid}=64$ for our experiments.

\vspace{4pt}
\noindent {\bf Comparison with State-of-the-art.}
Table~\ref{tab:all_main} shows the performance of our MoPEFT framework against the three most commonly used PEFT methods, i.e., LoRA \citep{hu2021lora}, Prefix Tuning (VPT Deep) \citep{jia2022vpt}, and Adapters \citep{he2021adapter}. We compare these methods against a vanilla SAM framework (Baseline), fully fine-tuning the SAM decoder on the target dataset (decoderFT), and 'simple' Visual Prompt Tuning (VPT) \citep{jia2022vpt}, which is similar to Prefix Tuning except that the tunable tensors are added to only the first Transformer block as opposed to all of them. We measure the Mean Intersection-over-Union (mIOU) to compare performance across all method and datasets.

\vspace{4pt}
\noindent {\bf Analysis of Gating Mechanism.}
The results in this section provide a better understanding of what the MoE learns during fine-tuning. To gain a better understanding of our gating mechanism, we conduct an analysis by tracking the frequency of the selection of each PEFT technique across different datasets during inference. We present our detailed results in Figure~\ref{fig:graphs}.

Most notable in our results is the fact that different datasets give more preference to different PEFT techniques. For instance, the graph depicting iSAID \citep{waqas2019isaid} (an \textit{Earth Monitoring} dataset in the MESS benchmark\citep{blumenstiel2024mess}), tends to select LoRA more often than the other two PEFT methods. Similarly, Kvasir-Instrument \citep{jha2021kvasir} (a \textit{Medical Imaging} dataset in the MESS benchmark \citep{blumenstiel2024mess}) tends to select Adapters more often, instead of LoRA or Prefix Tuning. This observation supports our initial claim that our gating mechanism learns to dynamically select appropriate PEFT techniques based on the provided data-task setup. This reinforces the significance of the MoPEFT framework in tailoring its selection to the unique characteristics of diverse datasets enhancing its effectiveness across different domains.









\vspace{-6pt}
\section{Conclusion}
\vspace{-5pt}
In this paper, introduce a new framework, MoPEFT, that is inspired by Mixture-of-Experts and dynamically learns to activate a particular PEFT technique based on a given data-task setup. 
We also present a comprehensive study of the top three PEFT techniques and compare their effectiveness with our proposed framework for fine-tuning the Segment Anything Model (SAM). Our results show that MoPEFT usually outperforms all traditional fine-tuning techniques on multiple datasets across different domains.

\section*{Acknowledgements}

This research was partly supported by the Air Force Office of Scientific Research (AFOSR) under SBIR grant FA9550-22-P-0009 with Intelligent Fusion Technology, AFOSR grant FA9550-20-1-0039, and the Empire State Development's Division of Science, Technology and Innovation through the University of Rochester Center of Excellence in Data Science. The authors would like to thank RIT Research Computing for making computing resources available for experimentation.
{
    \small
    \bibliographystyle{ieeenat_fullname}
    \bibliography{main}

\begin{thebibliography}{36}
\providecommand{\natexlab}[1]{#1}
\providecommand{\url}[1]{\texttt{#1}}
\expandafter\ifx\csname urlstyle\endcsname\relax
  \providecommand{\doi}[1]{doi: #1}\else
  \providecommand{\doi}{doi: \begingroup \urlstyle{rm}\Url}\fi

\bibitem[Achiam et~al.(2023)Achiam, Adler, Agarwal, Ahmad, Akkaya, Aleman, Almeida, Altenschmidt, Altman, Anadkat, et~al.]{achiam2023gpt}
Josh Achiam, Steven Adler, Sandhini Agarwal, Lama Ahmad, Ilge Akkaya, Florencia~Leoni Aleman, Diogo Almeida, Janko Altenschmidt, Sam Altman, Shyamal Anadkat, et~al.
\newblock Gpt-4 technical report.
\newblock \emph{arXiv preprint arXiv:2303.08774}, 2023.

\bibitem[Blumenstiel et~al.(2024)Blumenstiel, Jakubik, K{\"u}hne, and V{\"o}ssing]{blumenstiel2024mess}
Benedikt Blumenstiel, Johannes Jakubik, Hilde K{\"u}hne, and Michael V{\"o}ssing.
\newblock What a mess: Multi-domain evaluation of zero-shot semantic segmentation.
\newblock \emph{Advances in Neural Information Processing Systems}, 36, 2024.

\bibitem[Chowdhery et~al.(2023)Chowdhery, Narang, Devlin, Bosma, Mishra, Roberts, Barham, Chung, Sutton, Gehrmann, et~al.]{chowdhery2023palm}
Aakanksha Chowdhery, Sharan Narang, Jacob Devlin, Maarten Bosma, Gaurav Mishra, Adam Roberts, Paul Barham, Hyung~Won Chung, Charles Sutton, Sebastian Gehrmann, et~al.
\newblock Palm: Scaling language modeling with pathways.
\newblock \emph{Journal of Machine Learning Research}, 24\penalty0 (240):\penalty0 1--113, 2023.

\bibitem[He et~al.(2021)He, Liu, Ye, Tan, Ding, Cheng, Low, Bing, and Si]{he2021adapter}
Ruidan He, Linlin Liu, Hai Ye, Qingyu Tan, Bosheng Ding, Liying Cheng, Jia-Wei Low, Lidong Bing, and Luo Si.
\newblock On the effectiveness of adapter-based tuning for pretrained language model adaptation.
\newblock \emph{arXiv preprint arXiv:2106.03164}, 2021.

\bibitem[Hu et~al.(2021)Hu, Shen, Wallis, Allen-Zhu, Li, Wang, Wang, and Chen]{hu2021lora}
Edward~J Hu, Yelong Shen, Phillip Wallis, Zeyuan Allen-Zhu, Yuanzhi Li, Shean Wang, Lu Wang, and Weizhu Chen.
\newblock Lora: Low-rank adaptation of large language models.
\newblock \emph{arXiv preprint arXiv:2106.09685}, 2021.

\bibitem[ISPRS(2014)]{isprspostdam}
Isprs ISPRS.
\newblock 2d semantic labeling contest, 2014.

\bibitem[Jha et~al.(2021)Jha, Ali, Emanuelsen, Hicks, Thambawita, Garcia-Ceja, Riegler, de~Lange, Schmidt, Johansen, et~al.]{jha2021kvasir}
Debesh Jha, Sharib Ali, Krister Emanuelsen, Steven~A Hicks, Vajira Thambawita, Enrique Garcia-Ceja, Michael~A Riegler, Thomas de Lange, Peter~T Schmidt, H{\aa}vard~D Johansen, et~al.
\newblock Kvasir-instrument: Diagnostic and therapeutic tool segmentation dataset in gastrointestinal endoscopy.
\newblock In \emph{MultiMedia Modeling: 27th International Conference, MMM 2021, Prague, Czech Republic, June 22--24, 2021, Proceedings, Part II 27}, pages 218--229. Springer, 2021.

\bibitem[Jia et~al.(2022)Jia, Tang, Chen, Cardie, Belongie, Hariharan, and Lim]{jia2022vpt}
Menglin Jia, Luming Tang, Bor-Chun Chen, Claire Cardie, Serge Belongie, Bharath Hariharan, and Ser-Nam Lim.
\newblock Visual prompt tuning.
\newblock In \emph{European Conference on Computer Vision}, pages 709--727. Springer, 2022.

\bibitem[Ke et~al.(2024)Ke, Ye, Danelljan, Tai, Tang, Yu, et~al.]{ke2024segment}
Lei Ke, Mingqiao Ye, Martin Danelljan, Yu-Wing Tai, Chi-Keung Tang, Fisher Yu, et~al.
\newblock Segment anything in high quality.
\newblock \emph{Advances in Neural Information Processing Systems}, 36, 2024.

\bibitem[Kirillov et~al.(2023)Kirillov, Mintun, Ravi, Mao, Rolland, Gustafson, Xiao, Whitehead, Berg, Lo, et~al.]{kirillov2023segment}
Alexander Kirillov, Eric Mintun, Nikhila Ravi, Hanzi Mao, Chloe Rolland, Laura Gustafson, Tete Xiao, Spencer Whitehead, Alexander~C Berg, Wan-Yen Lo, et~al.
\newblock Segment anything.
\newblock In \emph{Proceedings of the IEEE/CVF International Conference on Computer Vision}, pages 4015--4026, 2023.

\bibitem[Lester et~al.(2021)Lester, Al-Rfou, and Constant]{lester2021prompt}
Brian Lester, Rami Al-Rfou, and Noah Constant.
\newblock The power of scale for parameter-efficient prompt tuning.
\newblock \emph{arXiv preprint arXiv:2104.08691}, 2021.

\bibitem[Li and Liang(2021)]{li2021prefix}
Xiang~Lisa Li and Percy Liang.
\newblock Prefix-tuning: Optimizing continuous prompts for generation.
\newblock \emph{arXiv preprint arXiv:2101.00190}, 2021.

\bibitem[Lin et~al.(2024)Lin, Tang, Ye, Cui, Zhu, Jin, Zhang, Ning, and Yuan]{lin2024moe}
Bin Lin, Zhenyu Tang, Yang Ye, Jiaxi Cui, Bin Zhu, Peng Jin, Junwu Zhang, Munan Ning, and Li Yuan.
\newblock Moe-llava: Mixture of experts for large vision-language models.
\newblock \emph{arXiv preprint arXiv:2401.15947}, 2024.

\bibitem[Liu et~al.(2019)Liu, Yao, Lu, Xie, and Li]{dram}
Yahui Liu, Jian Yao, Xiaohu Lu, Renping Xie, and Li Li.
\newblock Deepcrack: A deep hierarchical feature learning architecture for crack segmentation.
\newblock \emph{Neurocomputing}, 338:\penalty0 139--153, 2019.

\bibitem[Lyu et~al.(2020)Lyu, Vosselman, Xia, Yilmaz, and Yang]{lyu2020uavid}
Ye Lyu, George Vosselman, Gui-Song Xia, Alper Yilmaz, and Michael~Ying Yang.
\newblock Uavid: A semantic segmentation dataset for uav imagery.
\newblock \emph{ISPRS journal of photogrammetry and remote sensing}, 165:\penalty0 108--119, 2020.

\bibitem[Mahbod et~al.(2021)Mahbod, Schaefer, Bancher, L{\"o}w, Dorffner, Ecker, and Ellinger]{mahbod2021cryonuseg}
Amirreza Mahbod, Gerald Schaefer, Benjamin Bancher, Christine L{\"o}w, Georg Dorffner, Rupert Ecker, and Isabella Ellinger.
\newblock Cryonuseg: A dataset for nuclei instance segmentation of cryosectioned h\&e-stained histological images.
\newblock \emph{Computers in biology and medicine}, 132:\penalty0 104349, 2021.

\bibitem[Mateo-Garcia et~al.(2021)Mateo-Garcia, Veitch-Michaelis, Smith, Oprea, Schumann, Gal, Baydin, and Backes]{worldfloods}
Gonzalo Mateo-Garcia, Joshua Veitch-Michaelis, Lewis Smith, Silviu~Vlad Oprea, Guy Schumann, Yarin Gal, At{\i}l{\i}m~G{\"u}ne{\c{s}} Baydin, and Dietmar Backes.
\newblock Towards global flood mapping onboard low cost satellites with machine learning.
\newblock \emph{Scientific reports}, 11\penalty0 (1):\penalty0 7249, 2021.

\bibitem[Mazurowski et~al.(2023)Mazurowski, Dong, Gu, Yang, Konz, and Zhang]{mazurowski2023segment}
Maciej~A Mazurowski, Haoyu Dong, Hanxue Gu, Jichen Yang, Nicholas Konz, and Yixin Zhang.
\newblock Segment anything model for medical image analysis: an experimental study.
\newblock \emph{Medical Image Analysis}, 89:\penalty0 102918, 2023.

\bibitem[Neuhold et~al.(2017)Neuhold, Ollmann, Rota~Bulo, and Kontschieder]{mhpv1}
Gerhard Neuhold, Tobias Ollmann, Samuel Rota~Bulo, and Peter Kontschieder.
\newblock The mapillary vistas dataset for semantic understanding of street scenes.
\newblock In \emph{Proceedings of the IEEE international conference on computer vision}, pages 4990--4999, 2017.

\bibitem[Nguyen and Chamroukhi(2018)]{nguyen2018practical}
Hien~D Nguyen and Faicel Chamroukhi.
\newblock Practical and theoretical aspects of mixture-of-experts modeling: An overview.
\newblock \emph{Wiley Interdisciplinary Reviews: Data Mining and Knowledge Discovery}, 8\penalty0 (4):\penalty0 e1246, 2018.

\bibitem[Osco et~al.(2023)Osco, Wu, de~Lemos, Gon{\c{c}}alves, Ramos, Li, and Junior]{osco2023segment}
Lucas~Prado Osco, Qiusheng Wu, Eduardo~Lopes de Lemos, Wesley~Nunes Gon{\c{c}}alves, Ana Paula~Marques Ramos, Jonathan Li, and Jos{\'e}~Marcato Junior.
\newblock The segment anything model (sam) for remote sensing applications: From zero to one shot.
\newblock \emph{International Journal of Applied Earth Observation and Geoinformation}, 124:\penalty0 103540, 2023.

\bibitem[Paszke et~al.(2019)Paszke, Gross, Massa, Lerer, Bradbury, Chanan, Killeen, Lin, Gimelshein, Antiga, et~al.]{atlantis}
Adam Paszke, Sam Gross, Francisco Massa, Adam Lerer, James Bradbury, Gregory Chanan, Trevor Killeen, Zeming Lin, Natalia Gimelshein, Luca Antiga, et~al.
\newblock Pytorch: An imperative style, high-performance deep learning library.
\newblock \emph{Advances in neural information processing systems}, 32, 2019.

\bibitem[Pfeiffer et~al.(2020)Pfeiffer, R{\"u}ckl{\'e}, Poth, Kamath, Vuli{\'c}, Ruder, Cho, and Gurevych]{pfeiffer2020adapterhub}
Jonas Pfeiffer, Andreas R{\"u}ckl{\'e}, Clifton Poth, Aishwarya Kamath, Ivan Vuli{\'c}, Sebastian Ruder, Kyunghyun Cho, and Iryna Gurevych.
\newblock Adapterhub: A framework for adapting transformers.
\newblock \emph{arXiv preprint arXiv:2007.07779}, 2020.

\bibitem[Pieri et~al.(2024)Pieri, Mullappilly, Khan, Anwer, Khan, Baldwin, and Cholakkal]{pieri2024bimedix}
Sara Pieri, Sahal~Shaji Mullappilly, Fahad~Shahbaz Khan, Rao~Muhammad Anwer, Salman Khan, Timothy Baldwin, and Hisham Cholakkal.
\newblock Bimedix: Bilingual medical mixture of experts llm.
\newblock \emph{arXiv preprint arXiv:2402.13253}, 2024.

\bibitem[Radford et~al.(2021)Radford, Kim, Hallacy, Ramesh, Goh, Agarwal, Sastry, Askell, Mishkin, Clark, et~al.]{radford2021learning}
Alec Radford, Jong~Wook Kim, Chris Hallacy, Aditya Ramesh, Gabriel Goh, Sandhini Agarwal, Girish Sastry, Amanda Askell, Pamela Mishkin, Jack Clark, et~al.
\newblock Learning transferable visual models from natural language supervision.
\newblock In \emph{International conference on machine learning}, pages 8748--8763. PMLR, 2021.

\bibitem[Rahnemoonfar et~al.(2021)Rahnemoonfar, Chowdhury, Sarkar, Varshney, Yari, and Murphy]{rahnemoonfar2021floodnet}
Maryam Rahnemoonfar, Tashnim Chowdhury, Argho Sarkar, Debvrat Varshney, Masoud Yari, and Robin~Roberson Murphy.
\newblock Floodnet: A high resolution aerial imagery dataset for post flood scene understanding.
\newblock \emph{IEEE Access}, 9:\penalty0 89644--89654, 2021.

\bibitem[Sakaridis et~al.(2019{\natexlab{a}})Sakaridis, Dai, and Gool]{chasedb1}
Christos Sakaridis, Dengxin Dai, and Luc~Van Gool.
\newblock Guided curriculum model adaptation and uncertainty-aware evaluation for semantic nighttime image segmentation.
\newblock In \emph{Proceedings of the IEEE/CVF International Conference on Computer Vision}, pages 7374--7383, 2019{\natexlab{a}}.

\bibitem[Sakaridis et~al.(2019{\natexlab{b}})Sakaridis, Dai, and Gool]{darkzurich}
Christos Sakaridis, Dengxin Dai, and Luc~Van Gool.
\newblock Guided curriculum model adaptation and uncertainty-aware evaluation for semantic nighttime image segmentation.
\newblock In \emph{Proceedings of the IEEE/CVF International Conference on Computer Vision}, pages 7374--7383, 2019{\natexlab{b}}.

\bibitem[Seibold et~al.(2022)Seibold, Rei{\ss}, Sarfraz, Fink, Mayer, Sellner, Kim, Maier-Hein, Kleesiek, and Stiefelhagen]{paxray}
Constantin Seibold, Simon Rei{\ss}, Saquib Sarfraz, Matthias~A Fink, Victoria Mayer, Jan Sellner, Moon~Sung Kim, Klaus~H Maier-Hein, Jens Kleesiek, and Rainer Stiefelhagen.
\newblock Detailed annotations of chest x-rays via ct projection for report understanding.
\newblock \emph{arXiv preprint arXiv:2210.03416}, 2022.

\bibitem[Shazeer et~al.(2017)Shazeer, Mirhoseini, Maziarz, Davis, Le, Hinton, and Dean]{shazeer2017outrageously}
Noam Shazeer, Azalia Mirhoseini, Krzysztof Maziarz, Andy Davis, Quoc Le, Geoffrey Hinton, and Jeff Dean.
\newblock Outrageously large neural networks: The sparsely-gated mixture-of-experts layer, 2017.

\bibitem[Wang et~al.(2022)Wang, Mukherjee, Liu, Gao, Awadallah, and Gao]{wang2022adamix}
Yaqing Wang, Subhabrata Mukherjee, Xiaodong Liu, Jing Gao, Ahmed~Hassan Awadallah, and Jianfeng Gao.
\newblock Adamix: Mixture-of-adapter for parameter-efficient tuning of large language models.
\newblock \emph{arXiv preprint arXiv:2205.12410}, 1\penalty0 (2):\penalty0 4, 2022.

\bibitem[Waqas~Zamir et~al.(2019)Waqas~Zamir, Arora, Gupta, Khan, Sun, Shahbaz~Khan, Zhu, Shao, Xia, and Bai]{waqas2019isaid}
Syed Waqas~Zamir, Aditya Arora, Akshita Gupta, Salman Khan, Guolei Sun, Fahad Shahbaz~Khan, Fan Zhu, Ling Shao, Gui-Song Xia, and Xiang Bai.
\newblock isaid: A large-scale dataset for instance segmentation in aerial images.
\newblock In \emph{Proceedings of the IEEE/CVF Conference on Computer Vision and Pattern Recognition Workshops}, pages 28--37, 2019.

\bibitem[Wu et~al.(2021)Wu, Fu, Liu, Lim, Hoi, and Sun]{foodseg}
Xiongwei Wu, Xin Fu, Ying Liu, Ee-Peng Lim, Steven~CH Hoi, and Qianru Sun.
\newblock A large-scale benchmark for food image segmentation.
\newblock In \emph{Proceedings of the 29th ACM international conference on multimedia}, pages 506--515, 2021.

\bibitem[Xu et~al.(2021)Xu, Yen, Zhao, and Xiao]{xu2021rethinking}
Dongkuan Xu, Ian~EH Yen, Jinxi Zhao, and Zhibin Xiao.
\newblock Rethinking network pruning--under the pre-train and fine-tune paradigm.
\newblock \emph{arXiv preprint arXiv:2104.08682}, 2021.

\bibitem[Yang et~al.(2023)Yang, Gao, Li, Gao, Wang, and Zheng]{yang2023track}
Jinyu Yang, Mingqi Gao, Zhe Li, Shang Gao, Fangjing Wang, and Feng Zheng.
\newblock Track anything: Segment anything meets videos.
\newblock \emph{arXiv preprint arXiv:2304.11968}, 2023.

\bibitem[Yu et~al.(2020)Yu, Chen, Wang, Xian, Chen, Liu, Madhavan, and Darrell]{yu2020bdd100k}
Fisher Yu, Haofeng Chen, Xin Wang, Wenqi Xian, Yingying Chen, Fangchen Liu, Vashisht Madhavan, and Trevor Darrell.
\newblock Bdd100k: A diverse driving dataset for heterogeneous multitask learning.
\newblock In \emph{Proceedings of the IEEE/CVF conference on computer vision and pattern recognition}, pages 2636--2645, 2020.

\end{thebibliography}
}


\end{document}